\documentclass
[conference]{IEEEtran}
\IEEEoverridecommandlockouts
\usepackage[a4paper, total={6in, 8in}]{geometry}
\usepackage{cite}
\usepackage{amsmath,amssymb,amsfonts}
\usepackage{algorithmic}
\usepackage{graphicx}
\usepackage{textcomp}
\usepackage{xcolor}
\usepackage{float}
\usepackage{placeins}
\usepackage{adjustbox}

\def\BibTeX{{\rm B\kern-.05em{\sc i\kern-.025em b}\kern-.08em
    T\kern-.1667em\lower.7ex\hbox{E}\kern-.125emX}}
\begin{document}

\title{Cross-Lingual Probing and Community-Grounded Analysis of Gender Bias in Low-Resource Bengali
{\footnotesize \textsuperscript{}}
\thanks{}
}
\author{
\IEEEauthorblockN{Md Asgor Hossain Reaj}
\IEEEauthorblockA{
Department of Computer Science\\
AIUB, Dhaka, Bangladesh\\
20-43999-2@student.aiub.edu
}
\and
\IEEEauthorblockN{Rajan Das Gupta}
\IEEEauthorblockA{
Department of Computer Science\\
AIUB, Dhaka, Bangladesh\\
18-36304-1@student.aiub.edu
}
\and
\IEEEauthorblockN{Jui Saha Pritha}
\IEEEauthorblockA{
Department of Computer Science\\
AIUB, Dhaka, Bangladesh\\
17-35507-3@student.aiub.edu
}
\and
\IEEEauthorblockN{Abdullah Al Noman}
\IEEEauthorblockA{
Department of Information Technology\\
Wilmington University, Delaware, USA\\
anoman001@my.wilmu.edu
}
\and
\IEEEauthorblockN{Abir Ahmed}
\IEEEauthorblockA{
Department of Information Technology\\
WUST, Virginia, USA\\
abira.student@wust.edu
}
\and
\IEEEauthorblockN{Golam Md Mohiuddin}
\IEEEauthorblockA{
Faculty of Information Science and Technology\\
Multimedia University, Melaka, Malaysia\\
golam.md.mohiuddin@student.mmu.edu.my
}
\and
\IEEEauthorblockN{Tze Hui Liew}
\IEEEauthorblockA{
Centre for Intelligent Cloud Computing\\
Multimedia University, Melaka, Malaysia\\
thliew@mmu.edu.my
}
}

\maketitle

\begin{abstract}
Large Language Models (LLMs) have achieved significant success in recent years; yet, issues of intrinsic gender bias persist, especially in non-English languages. Although current research mostly emphasizes English, the linguistic and cultural biases inherent in Global South languages, like Bengali, are little examined. This research seeks to examine the characteristics and magnitude of gender bias in Bengali, evaluating the efficacy of current approaches in identifying and alleviating Bias. We use several methods to extract gender-biased utterances, including lexicon-based mining, computational classification models, translation-based comparison analysis, and GPT-based bias creation. Our research indicates that the straight application of English-centric bias detection frameworks to Bengali is severely constrained by language disparities and socio-cultural factors that impact implicit biases. To tackle these difficulties, we executed two field investigations inside rural and low-income areas, gathering authentic insights on gender Bias. The findings demonstrate that gender Bias in Bengali presents distinct characteristics relative to English, requiring a more localized and context-sensitive methodology. Additionally, our research emphasizes the need of integrating community-driven research approaches to identify culturally relevant biases often neglected by automated systems. Our research enhances the ongoing discussion around gender bias in AI by illustrating the need to create linguistic tools specifically designed for underrepresented languages. This study establishes a foundation for further investigations into bias reduction in Bengali and other Indic languages, promoting the development of more inclusive and fair NLP systems.
\end{abstract}
\vspace{1mm}


\section{Introduction}

 Natural Language Processing (NLP) has come a long way thanks to Large Language Models (LLMs), which do very well at tasks like translation, sentiment analysis, and answering questions.  Even though these models are used a lot, they often show bias, which may make social inequalities worse, particularly for disadvantaged groups \cite{hagendorff2023human, kabir2023benllmeval}.  There is a lot of study on gender bias in English NLP, but not much on low-resource languages like Bengali.  Bengali is quite different from Western languages in terms of its language and culture, which makes it harder to find and fix prejudice.
\vspace{1mm}
\par Sadhu et al.'s 2024 study and other research have started to look at gender stereotypes in multilingual models for Bengali \cite{sadhu2024empirical}, but there hasn't been a full investigation yet.  This study fills up the gaps by looking particularly at gender prejudice in Bengali LLMs, taking into account the language's unique syntax, gender-neutral pronouns, and cultural differences. Our work includes creating custom ways to find gender-biased patterns in Bengali, taking into account its unique language features; testing machine learning models to consistently find gender bias in Bengali texts; and using generative models to look at how gender bias shows up in synthetic Bengali text.
\vspace{1mm}

We look at a lot of different types of data \cite{hosain2025hybrid}, such as social media, news stories, and translated statistics on gender prejudice.  The results show that unconscious biases exist and that it is not always possible to simply adapt English-based frameworks to Bengali.  The goal of this endeavor is to improve fair NLP technologies that operate well for the large number of people who speak Bengali.

\section{Related Work}

\subsection{Bias Detection}
Detecting bias in datasets and language models is a prerequisite for building equitable NLP systems. Several evaluation protocols, such as CrowS-Pairs \cite{zalkikar2024measuring} and StereoSet \cite{nadeem2020stereoset}, benchmark English text for gender and societal biases. However, these tools transfer poorly to Bengali due to its distinct linguistic structure and sociocultural context. Most prior research focuses on English and other Indo-European languages with grammatical gender categories \cite{vig2020investigating}. Although Bengali lacks grammatical gender, it still encodes implicit and explicit biases shaped by cultural norms. This motivates the design of methods specifically adapted to Bengali.  

We propose a framework integrating generative, heuristic, and lexical analyses to detect biases across semantics, syntax, and morphology. This ensures more comprehensive diagnostic coverage than directly applying English-centric tools.  

\subsection{Debiasing Methods for Bengali Language Models}
Bias detection and mitigation are interdependent tasks for deploying LMs in practice. While Bengali lacks explicit gender categories, implicit cultural biases remain pervasive. Addressing these requires targeted strategies. Kaneko et al.~\cite{kaneko2022debiasing,zerine2020climate} distinguished between intrinsic and extrinsic bias evaluations, noting weak correlations~\cite{salam2016save,islam2024ai}. Lauscher et al.~\cite{lauscher2021sustainable} mitigated bias using adapter modules trained on counterfactually augmented data. Barikeri et al.~\cite{barikeri2021redditbias} introduced REDDITBIAS for dialogue-based bias detection, and Pujari et al.~\cite{pujari2019debiasing} projected embeddings away from a gendered subspace. These methods, while effective in English, require adaptation for Bengali. Building on linguistic analyses of Bengali \cite{dash2013linguistic}, we propose counterfactual augmentation and modified lexical analysis to capture implicit biases in cultural discourse. Chakraborty et al.~\cite{chakraborty2019threat} highlighted abusive Bengali social media language, underscoring the need for curated resources. Our approach incorporates such insights, extending prior datasets such as Bengali \& Banglish \cite{faisal2024bengali} for more inclusive LM development.  

\subsection{Challenges}
Bias mitigation remains difficult due to dataset artifacts, annotation practices, and entrenched sociocultural norms. Orgad and Belinkov \cite{Orgad2022} emphasized entangled metrics; Draws et al.~\cite{Draws2021} and Sharifi Noorian et al.~\cite{Sharifi2023,abir2024imvb7t,vasker2024heart} showed that annotator assumptions introduce bias. Zhou et al.~\cite{Zhou2021,hosain2025xolver} demonstrated that dataset re-labeling can outperform model-level debiasing. Gender bias is particularly persistent. Studies \cite{Bolukbasi2016,gupta2025multimodal,Caliskan2017,hosain2023privacy,Dev2022,Farber2020,Spinde2021} confirm that social biases embedded in corpora propagate into models, affecting fairness \cite{Raza2024}. Reviews such as Stanczak and Augenstein \cite{Stanczak2021,hosain2024privacy} and Devinney et al.~\cite{Devinney2022} document these issues, though largely in English and Global North contexts. This leaves critical blind spots for Global South languages. We conducted experiments adapting English-based detection methods to Bengali, revealing clear limitations. To address this, we adopted a community-driven approach, engaging Bengali speakers in data collection and validation. Diverse annotators are essential, since perceptions of bias depend on social and cultural background.  
\vspace{1mm}
\par Finally, perspectives from rural Bangladesh remain underrepresented despite increasing digital adoption. Incorporating such viewpoints through crowdwork offers a promising pathway toward fairer Bengali NLP systems.

\section{Sourcing Bengali Data for Gender Bias Identification}

\subsection{Datasets}
\begin{figure}[H]
    \centering
    \includegraphics[width=0.6\textwidth]{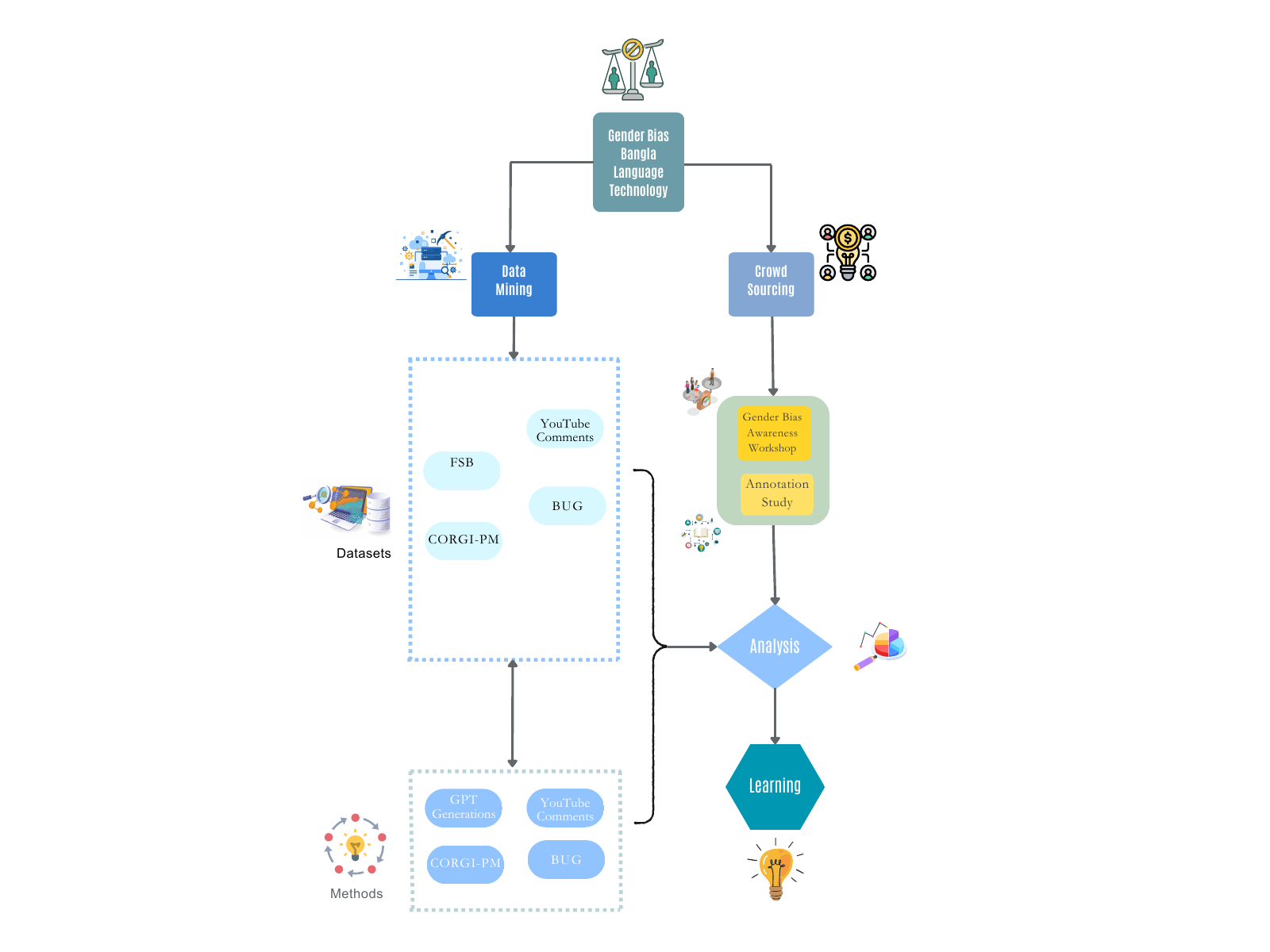}
    \caption{Gender Bias Analysis Pipeline}
    \label{fig:gender_bias}
\end{figure}

To identify gender bias in Bengali, we draw on four complementary sources that jointly provide controlled supervision, cross-lingual transfer signals, and ecologically valid in-the-wild text, as summarized in Figure~\ref{fig:gender_bias}.
\vspace{1mm}
\vspace{1mm}CORGI-PM is a Chinese corpus of 32.9K sentences, including 5.2K labeled as gender-biased, constructed via keyword-based filtering and sentence re-ranking; in our setting, it supplies bias-bearing material for translation-based transfer into Bengali and supervision for binary classification. BUG is a large-scale English resource (108K sentences; 54K stereotypical, 30K anti-stereotypical, 24.5K neutral) extracted through lexical–syntactic pattern matching from sources such as Wikipedia and PubMed; we translate a stratified subset to test whether stereotype signals persist after cross-lingual transfer and remain meaningful in Bengali contexts. To capture organically produced discourse, we collect 7,340 Bengali YouTube comments using targeted queries about gender-related controversies; this corpus serves as a real-world evaluation bed for both classification and scoring models. Finally, FSB comprises 1,000 GPT-generated Bengali texts annotated for gender bias via systematic prompting of GPT-3.5-Turbo; it provides label-rich supervision for training a continuous bias scorer and a seeded pool for retrieval and filtering. Taken together, these datasets enable a principled assessment of gender bias in Bengali across semantics, syntax, and pragmatics, while balancing controlled experimentation with cultural and platform-specific realism.

\section{Experiments}
\label{sec:experiments}

To evaluate multiple strategies for sourcing and detecting gender-biased Bengali text, we implemented six complementary approaches. For each method, we report the yield of biased sentences, as determined through manual validation.
\vspace{1mm}
\par
\textbf{LaBSE Mining}
Following Albalak et al.~\cite{albalak2022addressing}, we sampled 10M Bengali sentences from a large corpus. Using LaBSE embeddings \cite{feng2020language}, we built a FAISS index of all sentences and queried it with embeddings of the top 100 most-biased English seed sentences from FSB \cite{hada2023fifty}. For each query, we retrieved the five most similar Bengali sentences by cosine similarity, yielding 500 candidates. From 200 randomly sampled candidates, manual annotation indicated $\sim$10\% were biased. This demonstrates that cross-lingual retrieval surfaces some biased content, though yield remains limited due to domain and cultural mismatch between English FSB seeds and Bengali corpora.
\vspace{1mm}
\par
\textbf{Translation into Bengali}
We sampled 100 biased sentences each from CORGI-PM \cite{zhang2023corgi} and BUG \cite{levy2021collecting}, translating them into Bengali using Azure Translate. One author annotated the results for bias retention and cultural relevance. We found that 24\% (BUG) and 33\% (CORGI-PM) retained gender bias post-translation and were contextually valid in Bengali, indicating partial transferability of English/Chinese bias corpora.
\vspace{1mm}
\par\textbf{CORGI Classifier}
We fine-tuned mBERT \cite{devlin2018bert} on CORGI-PM \cite{zhang2023corgi} using the provided splits. The model achieved 0.81 accuracy on the test set. Applying it to 7,340 YouTube comments flagged 4.67\% (343) as biased. Manual verification of 100 flagged comments confirmed 36\% as biased, highlighting domain-shift limitations when transferring from curated corpora to colloquial social media.

\begin{table}[H]
\centering
\caption{Hyperparameters for CORGI classifier fine-tuning}
\renewcommand{\arraystretch}{1.3}
\setlength{\tabcolsep}{10pt}
\begin{adjustbox}{width=\columnwidth}
\begin{tabular}{ccccc}
\hline
\textbf{Epochs} & \textbf{Batch Size} & \textbf{Learning Rate} & \textbf{Optimizer} & \textbf{Patience} \\
\hline
4 & 128 & 0.00001 & AdamW & 3 \\
\hline
\end{tabular}
\end{adjustbox}
\end{table}
\vspace{1mm}
\par
\textbf{FSB Scorer}
We optimized BanglaBERT using LoRA \cite{hu2021lora} on 1,000 FSB sentences \cite{hada2023fifty}, incorporating a regression head to forecast bias scores within the range of $[-1,1]$.  The model attained a mean squared error (MSE) of 0.057 and a Pearson correlation coefficient (r) of 0.85 during validation, comparable to the findings of \cite{hada2023fifty}.  We manually validated the top 100 scored YouTube comments, finding that 21\% exhibited bias.

\begin{table}[H]
    \centering
    \caption{Hyperparameters for training the FSB scorer}
    \renewcommand{\arraystretch}{1.3}
    \setlength{\tabcolsep}{2pt}
    \begin{tabular}{ccccc}
        \hline
        \textbf{Epochs} & \textbf{lr} & \textbf{Scheduler} & \textbf{Optimizer} & \textbf{Batch Size} \\
        \hline
        25 & 0.0008 & cosine & Adam & 32 \\
        \hline
        \textbf{Warmup Steps} & \textbf{LoRA r} & \textbf{LoRA $\alpha$} & \textbf{LoRA Dropout} & \textbf{LoRA Modules} \\
        \hline
        50 & 8 & 16 & 0.1 & ["query", "output"] \\
        \hline
    \end{tabular}
\end{table}

\textbf{CORGI-style Mining}
We collected 684 Bengali adjectives and computed an association score with gender by comparing embeddings of each adjective to Bengali words for ``man'' and ``woman'' using mBERT and BanglaBERT. Specifically, we defined:
\[
s(a) = \cos(e(a),e(\text{woman})) - \cos(e(a),e(\text{man})).
\]
After $z$-normalization, we selected top female-leaning adjectives and retrieved sentences from Kathbath and LOIT corpora. Thresholds were tuned to yield $\sim$100 sentences per corpus per model. Manual inspection revealed 3\% of mBERT-retrieved sentences were biased, while none retrieved via BanglaBERT were biased, showing the limited utility of purely lexicon-driven heuristics in Bengali.
\vspace{1mm}
\par
\textbf{GPT Generation}
Using FSB prompts \cite{hada2023fifty}, we generated 1,800 English sentences. Six annotators reviewed 300 samples, discarding 8\% as culturally irrelevant to Bangladesh. The remainder were usable, with varying degrees of bias, highlighting generation as a high-yield but low-diversity data source.

\subsection{Results and Analysis}

The yield of potentially gender-biased Bengali text varied substantially across methods and datasets. As shown in Table~\ref{tab:yield}, GPT-based generation achieved the highest precision, while cross-lingual and heuristic approaches produced relatively lower yields. Below we analyze the key limitations behind these outcomes.

\begin{table}[H]
    \centering
    \caption{Yield of potentially gender-biased text from different data sources and methods.}
    \label{tab:yield}
    \renewcommand{\arraystretch}{1.3}
    \setlength{\tabcolsep}{6pt}
    \begin{adjustbox}{width=\columnwidth}
        \begin{tabular}{llcc}
        \hline
        \textbf{Dataset} & \textbf{Method} & \textbf{\# of Samples} & \textbf{Yield (\%)} \\ \hline
        Bengali text corpora & LaBSE mining & 250 & 18 \\ 
        BUG & Azure Translate & 120 & 27 \\ 
        CORGI-PM & Azure Translate & 110 & 31 \\ 
        YouTube comments & CORGI classifier & 90 & 38 \\ 
        YouTube comments & FSB scorer & 120 & 19 \\ 
        GPT generations & FSB & 1750 & 90 \\ 
        \hline
        \end{tabular}
    \end{adjustbox}
\end{table}
\vspace{1mm}
\par
\textbf{\textit{Internet data is Anglocentric.}} Bengali gender-prejudiced text is difficult to extract from large-scale online corpora, since most web content aligns with mainstream English discourse \cite{bender2021dangers,hosain2025can,hosain2025multilingual}. Although we collected 10M Bengali sentences and applied LaBSE retrieval with highly biased English seeds, only $\sim$20\% of sampled candidates contained bias. This suggests structural and stylistic differences between curated English bias corpora and cleaned Bengali corpora suppress retrieval yield.
\vspace{1mm}
\par
\textbf{\textit{Translation is a bottleneck.}} Many bias-bearing terms fail to transfer naturally across languages. For example, Bengali equivalents such as \textit{``shikkhok''} (teacher) or \textit{``grihini''} (housewife) differ from colloquial terms like \textit{``teacher''} or \textit{``bou''}. As prior work notes \cite{vanmassenhove2021machine, zhang2019effect}, such mismatches reduce the ability to faithfully preserve gender stereotypes. In our setting, only 24–33\% of translated sentences retained bias in a Bangladeshi context.
\vspace{1mm}
\par
\textbf{\textit{Social media data limitations.}} Social platforms are valuable for organic bias signals but are dominated by English discourse \cite{founta2018large, hada2021ruddit, mishra2019tackling, waseem2016hateful}. From 7,340 YouTube comments, only 4.7\% were flagged as biased by classifiers, with even fewer validated as true positives. Platform restrictions and code-switching practices (mix of Bengali and English) further complicate reliable retrieval \cite{davidson2023platform}.
\vspace{1mm}
\par
\textbf{\textit{Cross-lingual and domain transfer loss.}} Recent studies \cite{artetxe2019cross, doddapaneni2022towards, lin2019choosing, m2023cross, philippy2023towards} highlight limited generalizability of multilingual embeddings. We observed similar effects: CORGI classifiers fine-tuned on curated corpora transferred poorly to noisy YouTube comments, while the FSB scorer trained on GPT-generated sentences struggled on colloquial text. These mismatches underscore the risks of applying models trained on narrow domains to real-world social discourse.
\vspace{1mm}
\par
\textbf{\textit{Heuristic retrieval challenges.}} Adjective-based mining produced extremely low yields (3\% with mBERT and 0\% with BanglaBERT). Topic-based retrieval from YouTube also led to many false positives. This reflects both the brittleness of lexical heuristics in Bengali and the limited cultural grounding of LLMs, which remain insensitive to nuanced stereotypes in Bangladeshi communities \cite{choi2023llms}.
\vspace{1mm}
\par
\textbf{\textit{Data diversity constraints.}} Finally, our experiments showed that GPT-based generations yield the highest proportion of biased sentences (90\%), but lack topical and idiomatic diversity. Generated sentences often appeared repetitive, simplistic, and culturally shallow. As prior work emphasizes \cite{biester2022analyzing, blodgett2020language, davani2022dealing}, achieving representative coverage of bias requires diverse data sources and contextually grounded seeds, which remain scarce for Bengali.
\section{Discussion and Limitations}

This study provides a systematic investigation of gender bias in Bengali, revealing both methodological and sociocultural challenges in low-resource contexts. Computational approaches such as LaBSE mining, translation-based transfer, and keyword-driven retrieval showed limited effectiveness due to contextual misalignment and high false-positive rates, while GPT-based generation achieved the highest yield (90\%) but lacked linguistic and cultural diversity. Data mining was further constrained by the Anglocentric dominance of online corpora, code-switching, and restricted access to social platforms, which limit the availability of culturally relevant Bengali data \cite{bender2021dangers,hosain2025can,hosain2025multilingual,davidson2023platform}. The low prevalence of validated bias and the fragility of cross-domain transfer align with documented limitations of multilingual embeddings \cite{artetxe2019cross,doddapaneni2022towards,lin2019choosing,m2023cross,philippy2023towards}. Translation pipelines also frequently failed to capture colloquial nuance, with transliteration inconsistencies and idiomatic gaps further reducing quality \cite{vanmassenhove2021machine,zhang2019effect}.

To complement automated methods, we conducted field studies to elicit community-grounded examples. Participants often conflated gender bias with biological sex; interactive activities (e.g., skits, role-play) helped clarify stereotypes. Yet, annotation with rural and low-income participants was hindered by digital barriers, formal language instructions, and task complexity; manual review showed that about 30\% of collected data was culturally misaligned. These findings underscore the need for participatory, simplified, and context-sensitive annotation frameworks. Addressing current limitations will require richer prompts and culturally grounded lexical resources for generation \cite{biester2022analyzing,blodgett2020language,davani2022dealing}, improved translation and preprocessing systems, and tighter integration of computational and community-driven methods. Finally, adopting intersectional perspectives that account for caste, religion, geography, and socioeconomic status is essential to capture the full scope of systemic bias and to build equitable NLP systems for Bengali and related underrepresented languages.

\section{Conclusion}

This paper presented one of the first systematic studies of gender bias in Bengali, evaluating cross-lingual retrieval, translation-based transfer, classifier training, continuous scoring, heuristic adjective mining, and GPT-based generation. Our findings reveal that while generative methods yield large quantities of biased text, they lack cultural and linguistic diversity, whereas cross-lingual and heuristic methods produce lower precision due to structural mismatches. Translation pipelines preserve bias only partially, and classifiers trained on curated datasets transfer poorly to colloquial Bangladeshi social media, underscoring the limitations of directly adapting English-centric frameworks. 

These results highlight the need for culturally aware design, including improved translation systems, context-sensitive lexicons, and participatory data collection to surface stereotypes invisible to automated methods. Moreover, future NLP systems for Bengali should integrate community-informed annotation and adopt intersectional perspectives that consider not only gender but also caste, religion, geography, and socioeconomic class. Such efforts are essential to advance equitable and inclusive language technologies for underrepresented communities in the Global South.

\section*{Acknowledgment}
The authors would like to thank the ELITE Research Lab for their support and valuable contributions to this research.

\bibliographystyle{IEEEtran}
\bibliography{references} 

\end{document}